\def\year{2022}\relax
\documentclass[letterpaper]{article} 
\usepackage{aaai22}  
\usepackage{times}  
\usepackage{helvet}  
\usepackage{courier}  
\usepackage[hyphens]{url}  
\usepackage{graphicx} 
\usepackage{natbib}  
\usepackage{caption} 
\DeclareCaptionStyle{ruled}{labelfont=normalfont,labelsep=colon,strut=off} 
\usepackage{bm}
\usepackage{amsmath}
\usepackage{amssymb}
\usepackage{booktabs}
\usepackage{threeparttable}
\usepackage{multicol}
\usepackage{multirow}

\usepackage[ruled]{algorithm2e}
\title{Multi-Scale Dynamic Coding Improved Spiking Actor Network \\for Reinforcement Learning}
\author{
    Duzhen Zhang\textsuperscript{\rm 1,2}\equalcontrib,
    Tielin Zhang\textsuperscript{\rm 1,2}\equalcontrib\footnote{Corresponding author.},
    Shuncheng Jia\textsuperscript{\rm 1,2},
    Bo Xu\textsuperscript{\rm 1,2,3}\footnotemark[2]
}
\affiliations{
    \textsuperscript{\rm 1}Research Center for Brain-inspired Intelligence, Institute of Automation, Chinese Academy of Sciences, Beijing, China\\
    \textsuperscript{\rm 2}School of Artificial Intelligence, University of Chinese Academy of Sciences, Beijing, China\\
    \textsuperscript{\rm 3}Center for Excellence in Brain Science and Intelligence Technology, Chinese Academy of Sciences, Shanghai, China\\

    $\{$xubo, tielin.zhang, zhangduzhen2019$\}$@ia.ac.cn
}

\usepackage{bibentry}

\begin{document}

\maketitle

\begin{abstract}
With the help of deep neural networks (DNNs), deep reinforcement learning (DRL) has achieved great success on many complex tasks, from games to robotic control. Compared to DNNs with partial brain-inspired structures and functions, spiking neural networks (SNNs) consider more biological features, including spiking neurons with complex dynamics and learning paradigms with biologically plausible plasticity principles. Inspired by the efficient computation of cell assembly in the biological brain, whereby memory-based coding is much more complex than readout, we propose a multiscale dynamic coding improved spiking actor network (MDC-SAN) for reinforcement learning to achieve effective decision-making. The population coding at the network scale is integrated with the dynamic neurons coding (containing 2nd-order neuronal dynamics) at the neuron scale towards a powerful spatial-temporal state representation. Extensive experimental results show that our MDC-SAN performs better than its counterpart deep actor network (based on DNNs) on four continuous control tasks from OpenAI gym. We think this is a significant attempt to improve SNNs from the perspective of efficient coding towards effective decision-making, just like that in biological networks.
\end{abstract}

\section{Introduction}
Reinforcement learning (RL) is staying at an increasingly important position in the research area of machine learning \cite{kaelbling1996reinforcement}, where agents interact with the environment in a trial-and-error manner and learn an optimal policy by maximizing accumulated rewards to reach excellent decision-making performance \cite{sutton2018reinforcement}. However, it is a general but challenging problem for all conventional RL algorithms to extract features from complex state space efficiently. With deep neural networks (DNNs) as powerful function approximators, deep reinforcement learning (DRL) has resolved this problem to some extent by directly learning a mapping from raw state space to action space and has been well applied on various applications, including recommendation systems~\cite{zou2019reinforcement,warlop2018fighting}, games~\cite{mnih2015human,vinyals2019grandmaster,DBLP:journals/corr/abs-2201-06257}, and robotic control~\cite{duan2016benchmarking,lillicrap2016continuous}, etc.

However, the powerful DRL is still far from efficient reward-based learning in the biological brain, where spiking neurons with more complex dynamics and learning paradigms with biologically plausible plasticity principles are integrated to generate complex cognitive functions. The biological brain makes efficient computation possible by cell assembly \cite{RN1047} which focuses more on spatial-temporal coding for memory than readout for decision-making. Compared to DNNs, SNNs have more significant potential in simulating brain-inspired topology and functions due to their complex dynamics.
For instance, SNNs can be seamlessly compatible with multiscale dynamic coding, including network and neuron scales, towards a powerful temporal information representation.
The SNNs inherently transmit and compute information with dynamic spikes distributed over time~\cite{maass1997networks}. Further research on them might help us open the black box of efficient information coding of the brain \cite{painkras2013spinnaker}. Hence, we think RL using SNNs may be better than using DNNs.

To this end, we propose a multiscale dynamic coding improved spiking actor network (MDC-SAN) to simulate the cell assembly in the biological brain, which contains a complex spiking coding module for state representation but a simple readout module for action inference. The coding module combines population coding and dynamic neurons (DNs) coding, making it more potent on state representation at both network and neuron scales. Specifically, for a given input state, we encode each dimension in individual neuron populations with learnable receptive fields. Then the coded analog information is directly delivered to the network as input. Inside the network, we propose novel DNs to improve SNNs for a better information representation during spatial-temporal learning. The DNs contain 2nd-order dynamics of membrane potentials supported by key dynamic parameters. These parameters are self-learned from one of OpenAI gym \cite{brockman2016openai} tasks (e.g., Ant-v3) first, and then extend to other similar tasks (e.g., HalfCheetah-v3, Walker2d-v3, and Hopper-v3). 
After dynamic coding, we average the accumulated spikes in a predefined time window to obtain the average firing rate, further used to infer the output action by a simple readout module.

For effective learning, the proposed MDC-SAN is trained in conjunction with deep critic networks using the Twin Delayed Deep Deterministic policy gradient (TD3) algorithm \cite{fujimoto2018addressing,tang2020reinforcement}.
We evaluate the trained MDC-SAN on four standard OpenAI gym tasks \cite{brockman2016openai}, including Ant-v3, HalfCheetah-v3, Walker2d-v3, and Hopper-v3. Experimental results demonstrate that multiscale dynamic codings, including population coding and complex spatial-temporal coding of DNs, are consistently beneficial to the performance of the MDC-SAN. 
In addition, the proposed MDC-SAN significantly outperforms its counterpart deep actor network (DAN) on the above four tasks under the same experimental configurations.

The main contributions of this paper can be summarized as follows:

\begin{itemize}
\item We propose a MDC-SAN to simulate the cell assembly in the biological brain for effective decision-making, which contains a complicated coding module for state representation from network scale and neuron scale, and a simple readout module for action inference.

\item For the network scale, we apply population coding to increase the representation capacity of the network, which encodes each dimension of the input state in individual neuron populations with learnable receptive fields. We also have verified the advantages of population coding and comprehensively compared the impact of various input coding methods on performance.

\item For the neuron scale, we construct novel DNs, which contain 2nd-order neuronal dynamics for complex spatial-temporal information coding. We also analyze the membrane-potential dynamics of DNs and demonstrate the performance advantage of DNs against standard leaky-integrate-and-fire (LIF) neurons. 

\item  Under the same experimental configurations, our MDC-SAN, which integrates population coding and DNs coding, achieves better performance than its counterpart DAN in each task. To the best of our knowledge, our work is the first to achieve state-of-the-art performance on multiple continuous control tasks with SNNs.

\end{itemize}

\section{Related Work}
\paragraph{Integrating SNNs with RL} Recently, the literature has grown up around introducing SNNs into RL algorithms~\cite{florian2007reinforcement,o2013spiking,yuan2019reinforcement,doya2000reinforcement,fremaux2013reinforcement}. 
These approaches are typically based on reward-modulated local plasticity rules that perform well in simple control tasks but commonly fail in complex robotic control tasks due to limited optimization capability.

To address the limitation, some approaches integrate SNNs with DRL optimization. One of the approaches directly converts Deep Q-Networks (DQNs)~\cite{mnih2015human} to SNNs and achieves competitive scores on Atari games with discrete action space \cite{patel2019improved,tan2021strategy}. However, these converted SNNs usually exhibit inferior performance to DNNs with the same structure \cite{rathi2019enabling}. Another approach is based on a hybrid learning framework. It achieves success in the mapless navigation task of a mobile robot, where the SAN is trained in conjunction with deep critic networks using a DRL algorithm \cite{tang2020reinforcement}. We also want to highlight this hybrid viewpoint during training and further extend it with multiscale dynamic codings that have played vital roles in the efficient information representation of SNNs.

\paragraph{Information coding methods in SNNs} There are two categories of the input coding scheme in SNNs, rate coding, and temporal coding. Rate coding uses the firing rate of spike trains in a time window to encode information, where input real numbers are converted into spike trains with a frequency proportional to the input value~\cite{cheng2020lisnn}. Temporal coding encodes information with the relative timing of individual spikes, where input values are usually converted into spike trains with the precise time~\cite{comsa2020temporal,sboev2018spiking}. Besides that, population coding is special in integrating these two types. For example, each neuron in a population can generate spike trains with precise time and also contain a relation with other neurons (e.g., Gaussian receptive field) for better information encoding at a global scale \cite{georgopoulos1986neuronal}.

For the neuron coding scheme in SNNs, there are various types of spiking neurons~\cite{tuckwell1988introduction}. The integrate-and-fire (IF) neuron is the simplest neuron type. It fires when the membrane potential exceeds the firing threshold, and the potential is then reset as a predefined resting membrane potential~\cite{rathi2020diet}. Another leaky integrate-and-fire (LIF) neuron allows the membrane potential to keep shrinking over time by introducing a leak factor~\cite{gerstner2002spiking}. They are commonly used as standard 1st-order neurons. Moreover, the Izhikevich neuron with 2nd-order equations of membrane potential is proposed, which can better represent the complex neuron dynamics, but requires some predefined hyper-parameters~\cite{izhikevich2003simple}.

\begin{figure*}[t]
\centering
\includegraphics[width=0.85\textwidth]{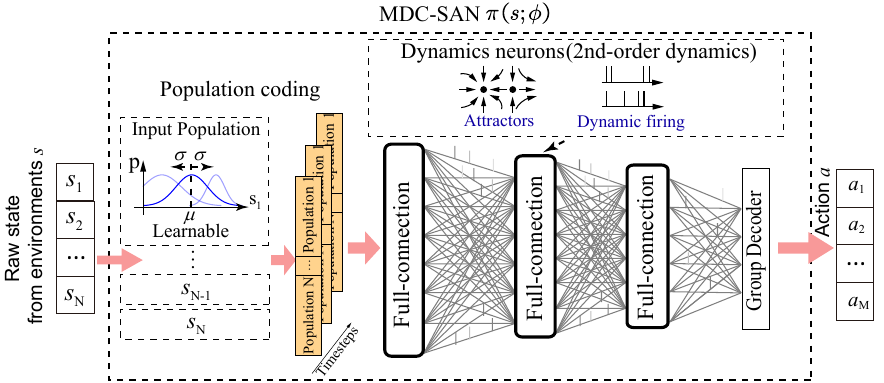} 
\caption{The overall architecture of proposed MDC-SAN.}
\label{fig-PDSAN}
\end{figure*}

\section{Methods}
The overview of our MDC-SAN is presented in Figure \ref{fig-PDSAN}, which contains an efficient coding module for spatial-temporal state representation and a relatively simple readout module for action inference. The MDC-SAN simulates the cell assembly in the biological brain by incorporating both population coding at the network scale and dynamic neurons coding at the neuron scale. For the population coding, each dimension of the input state is encoded with a group of dynamic receptive fields first and then fed into the SAN. For the dynamic neurons coding, the DNs inside of the SAN contain 2nd-order dynamic membrane potentials with up to two equilibrium points to describe complex neuronal dynamics. Finally, the average firing rate of the accumulated spikes in a predefined time window is decoded into output action with an additional group decoder.

\subsection{Population Coding} \label{pop-coding}

For a input state $\bm{s}\in \mathbb{R}^N$, it is encoded as an input $\bm{I}_t, t=\{1,2,...,T_1\}$ for each time step, where $T_1$ is the time window of the SNNs.

\paragraph{Pure population coding $C_{pop}$}

We create a population of neurons $P_{i}$ to encode each dimension of state $s_{i}$, where each neuron $P_{i,j}$ in the population has a Gaussian receptive field $(\mu_{i,j},\sigma_{i,j})$ with two learnable parameters of mean and standard deviation. The population coding $C_{pop}$ is formulated as:

\begin{equation}
\left\{\begin{array}{l}
A_{P_{i,j}}=\exp^{-\frac{(s_i-\mu_{i,j})^2}{2\sigma_{i,j}^2}} \\
\bm{A}_{P} = \left [A_{P_{1,1}},\dots,A_{P_{i,j}},\dots,A_{P_{N,J}} \right ]\\
\bm{I}_t = \bm{A}_{P}\\
\end{array}\right.\text{,}
\end{equation}

where $i$ is index of the input state ($i=1,...,N$), $j$ is index of neurons in a population ($j=1,...,J$), $\bm{A}_{P}$ is the stimulation strength after population coding, used as network input $\bm{I}_t$ directly.

There are other candidate input coding methods that combine population coding and rate coding (including uniform coding, Poisson coding, and deterministic coding). They contain two phases \cite{tang2020deep}: first the state $\bm{s}$ is transformed into the stimulation strength $\bm{A}_{P}$ by population coding and then the computed $\bm{A}_{P}$ is used to generate the input $\bm{I}_t$ by rate coding. We formalize these methods as follows.

\paragraph{The population and uniform coding ($C_{pop}$+$C_{uni}$)}

We generate random numbers $Rand_k(t)$ from 0 to 1 evenly distributed, which has the same size as the input stimulation strength $\bm{A}_{P}$ at every time step. Then we compare every generated random number with its corresponding input data. If the generated random number is less than its input data, $I_{k,t}$ is set to 1. Otherwise, it is set to 0, formulated as:
\begin{equation}
I_{k,t} = \begin{cases}
1, & A_{P_{k}} > Rand_k(t); \\
0,& \text{Otherwise.}
\end{cases}\text{,}
\end{equation}
where $k$ is the index of input stimulation strength ($k=1,...,N*J$).

\paragraph{The population and Poisson coding ($C_{pop}$+$C_{poi}$)}

Considering that the Poisson process can be considered as the limit of a Bernoulli process, the input stimulation strength $\bm{A}_{P}$ containing probability can be used for drawing the binary random number. The $I_{k,t}$ will draw a value 1 according to the $k^{th}$ probability value $A_{P_{k}}$ given in $\bm{A}_{P}$, formulated as:
\begin{equation}
P(I_{k,t}=1) = C_R^r A_{P_{k}}^r\left(1-A_{P_{k}}\right)^{R-r} \\
\text{,}
\end{equation}

\paragraph{The population and deterministic coding ($C_{pop}$+$C_{det}$)}

The input stimulation strength $\bm{A}_{P}$ acts as the presynaptic inputs to the postsynaptic neurons~\cite{tang2020deep}, formulated as:
\begin{equation}
V_{k,t} = V_{k,t-1} + A_{P_{k}}
\end{equation}

\begin{equation}
I_{k,t} =
\left\{\begin{array}{l}
\begin{matrix}
     1  & If (V_{k,t}>1)\\
     0  & Else
\end{matrix}
\end{array}\right.\text{,}
\end{equation}
where $V_{k,t}$ is reset as $V_{k,t} - 1$ when $I_{k,t}=1$, $V_{k,t}$ is pseudo membrane voltage.


\subsection{Dynamic Neurons}

This section introduces the traditional 1st-order neurons (e.g., LIF neurons) with a maximum of one equilibrium point and then defines the improved 2nd-order DNs with up to two equilibrium points. The procedure for constructing DNs is also introduced in the following sections.

\paragraph{The traditional 1st-order neurons}

The traditional 1st-order neurons in SNNs are the LIF neurons, which are the simplest abstraction of the Hodgkin–Huxley model. To show the essential equilibrium point characteristics of LIF neurons, here we give a simple definition of LIF neurons as the following description:
\begin{equation}
\tau\frac{dV_{i,t}}{dt} = -V_{i,t}+In_t \text{.}
\end{equation}
where $V_{i,t}$ is dynamic membrane potential for neuron $i$ at time $t$, $In_t$ is input represented as integrated post-synaptic potential. The single equilibrium point can be calculated as $V^*_{i,t}$ with input $In_t$ within period time of $\tau$.

\begin{figure}[t]
\centering
\includegraphics[width=0.9\columnwidth]{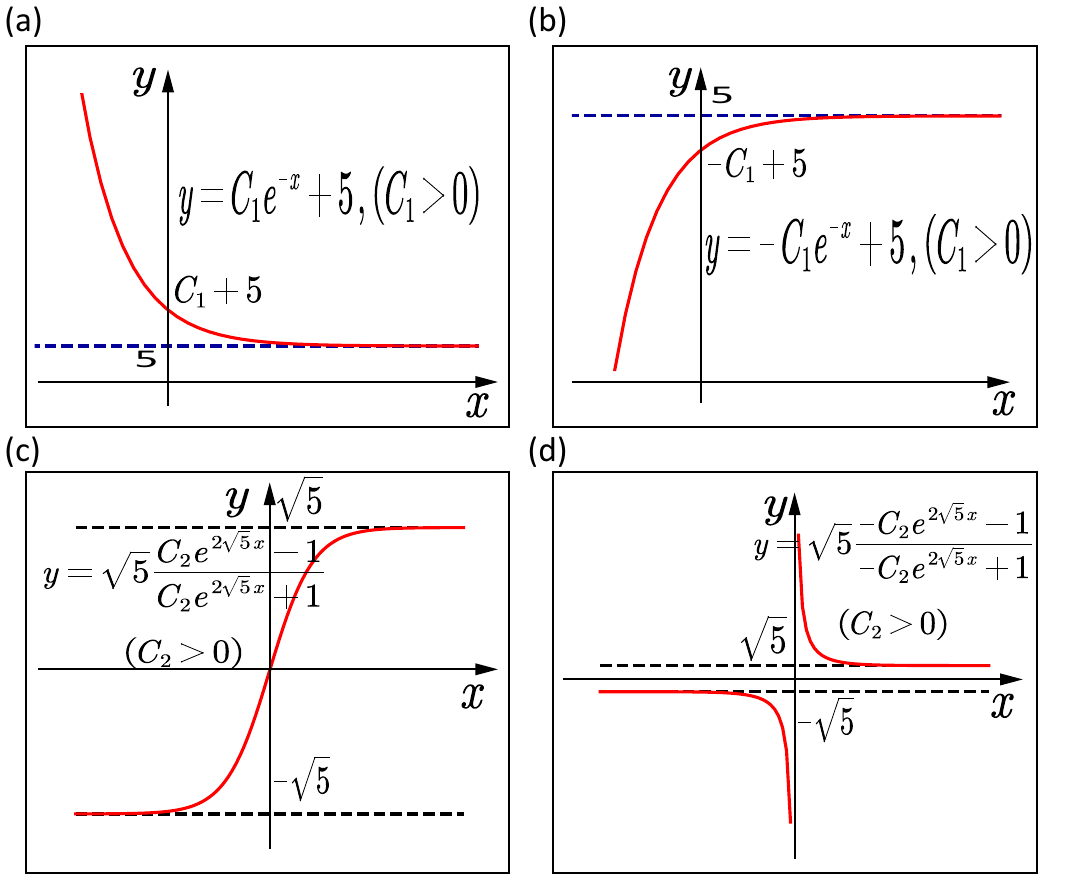} 
\caption{Dynamics of membrane potentials with different equilibrium points.}.
\label{fig-field}
\end{figure}

The number of equilibrium points will be the key to distinguishing different neuronal dynamics levels. For example, the dynamic field of $V_{i,t}$ for LIF model is reaching the single attractor $V^*_{i,t}=5$. When the firing threshold is bigger than $V^*_{i,t}$, the neuron will be mostly leaky (Fig. \ref{fig-field}(a)), or else it will be continuously firing (Fig. \ref{fig-field}(b)).

\paragraph{The designed 2nd-order DNs}

The neurons with higher-order dynamics mean that these neurons' number of equilibrium points will be more than one. Here we set it as 2 for simplicity. The 2nd-order neuronal dynamics is shown as follows:

\begin{equation}
\frac{dV_{j,t}}{dt}=V_{j,t}^2 - V_{j,t} - U_{j,t}+ In_{i,t} \text{,}
\label{equa_2order_V}
\end{equation}
where $V_{j,t}^2$ and $V_{j,t}$ are membrane potentials with different degrees of dynamics, $U_{j,t}$ is a resistance item for simulating hyper-polarization. The dynamic membrane potential $V_{j,t}$ will be attracted or non-stable at some points when we set the ordinary differential equation as 0. Fig. \ref{fig-field} (c) and Fig. \ref{fig-field} (d) show a diagram depicting dynamic fields of membrane potential with $N=2$, where the period for reaching stable points take around time $\tau$. For simplicity, we formulate traditional $mV$ or $ms$ units as 1. Besides membrane potential, some other implicit variables are also used for the description of 2nd-order dynamics, shown as follows:

\begin{equation}
\begin{cases}
\frac{dU_{j,t}}{dt}=\theta_{v} V_{j,t} - \theta_{u} U_{j,t} \\
V_{j,t}=\theta_r & if(V_{j,t}>V_{th})\\
U_{j,t}=U_{j,t}+\theta_s & if(spike)
\end{cases}\text{,}
\label{equa-MDN}
\end{equation}
where equilibrium point of $V_{j,t}$ is decided by both $U_{j,t}$ and input currents $In_t$. The number and value of equilibrium points will be dynamically affected by four parameters of $\theta_{v}$ (conductivity of $V$), $\theta_{u}$ (conductivity of $U$), $\theta_{r}$ (reset membrane potential), and $\theta_{s}$ (spike improvement of $U$), which is different from Izhikevich neurons \cite{izhikevich2003simple} by only focusing on higher-order dynamics.

\begin{algorithm}[t]
\caption{Forward propagation in MDC-SAN}
\label{algo1}
Initialize coding means $\bm{\mu}$ and standard deviations $\bm{\sigma}$ for all population encoders\;
Randomly initialize synaptic weight $\bm{W}$ and biases $\bm{b}$ for each SNN layer\;
Load the best dynamic parameters of DNs $\bm{\theta}^* = (\theta_v^*,\theta_u^*,\theta_r^*,\theta_s^*)$ (pre-learning from a task)\;
Randomly initialize decoding weight vectors $\bm{W}^d$ and bias $b^d$ for each action dimension\;
Initialize the current decay factor $d_c$ and firing threshold $V_{th}$\;
$N-$dimensional input state, $\bm{s}$\;
Inputs from populations generated by the input coding module: $\bm{A}_P$\;

\For{$t=1,...,T_{1}$}{
    Inputs at timestep t: $\bm{O}_t^{(0)}=\bm{I}_t$=$\bm{A}_P$\;
    \For{l=1,...,L}{
      Update DNs in layer $l$ at timestep $t$ based on spikes from layer $l-1$:\\
      $\bm{C}^{(l)}_t = d_c \cdot \bm{C}^{(l)}_{t-1} + \bm{W}^{(l)}\bm{O}^{(l-1)}_t +\bm{b}^{(l)}$\;
      $\bm{V}^{(l)}_t = \bm{V}^{(l)}_{t-1}\cdot (1-\bm{O}^{(l)}_{t-1}) + \bm{O}^{(l)}_{t-1}\cdot \theta_r^*$\;
      $\bm{U}^{(l)}_t = \bm{U}^{(l)}_{t-1} + \bm{O}^{(l)}_{t-1}\cdot \theta_s^*$\;
      $\bm{V}_{delta} = {\bm{V}^{(l)}_t}^2 - \bm{V}^{(l)}_t-\bm{U}^{(l)}_t+\bm{C}^{(l)}_t$\;
      $\bm{U}_{delta}=\theta_v^*\cdot\bm{V}^{(l)}_t-\theta_u^*\cdot\bm{U}^{(l)}_t$\;
      $\bm{V}^{(l)}_t=\bm{V}^{(l)}_t+V_{delta}$\;
      $\bm{U}^{(l)}_t=\bm{U}^{(l)}_t+U_{delta}$\;
      $\bm{O}^{(l)}_t=(\bm{V}^{(l)}_t>V_{th})$\;
    }
}
Sum up the output spikes: $\bm{sc}=\sum_{t=1}^{T_1}\bm{O}^{(L)}_t$\;
Compute the average firing rate: $\bm{fr} = \bm{sc}/T_1$\;
Divide $\bm{fr}$ into $M$ output groups:$\{\bm{fr}_{m}\},m=1,...,M$\;
Generate $M$-$ $dimensional action $\bm{a}$ by a grouped decoder: $a_m = \bm{W}^{d}_m\cdot\bm{fr}_{m}+b_{m}^d,m=1,...,M$\;
\end{algorithm}

\paragraph{The procedure for constructing the DNs}

The construction of DNs is mainly based on identifying some key parameters in them. As $\theta_{v,u,r,s}$, for example, each set of these four dynamic parameters describes a dynamic state of a spiking neuron. Hence, we want to obtain a set of optimal dynamic parameters.

We randomly initialize the dynamic parameters ($\theta_{v,u,r,s}$) of each neuron in the network. Together with synaptic weights, these dynamic parameters will be tuned on one of the tasks with the TD3 algorithm. After learning, where most of the learnable variables have reached stable points, these parameters will be plotted and clustered with the k-means method to get the best center of $\theta_{v,u,r,s}$ parameters. The four key parameters corresponding to the best center will be further used as the fixed configuration of all dynamic neurons for all tasks.

\subsection{The Simple Readout Module}

The spikes at the output layer are summed up in a predefined time window to compute the average firing rate first. Then, the output action $\bm{a}$ is returned as the weighted sum of the computed average firing rate by a grouped decoder. More details can be found in the forward propagation of MDC-SAN (Algorithm \ref{algo1}).


\subsection{The Learning of MDC-SAN with TD3}
The MDC-SAN is trained in conjunction with deep critic networks (i.e., a multi-layer fully-connected network) using the TD3 algorithm \cite{fujimoto2018addressing,tang2020reinforcement}. During training, the MDC-SAN infers an action $\bm{a}$ from a given state $\bm{s}$ to represent the policy, and a deep critic network estimates the associated action-value $\mathit{Q(s,a)}$ to guide the MDC-SAN to learn a better policy. We evaluate the trained MDC-SAN on a suit of continuous control benchmarks and compare its performance with its counterpart DAN (i.e., a multi-layer fully-connected network) under the same settings. The specific training procedure of the TD3 algorithm can be found in \cite{fujimoto2018addressing}.

\subsection{Tuning MDC-SAN with Approximate BP}

It is a challenge to tune parameters well in a network at multi scales, e.g., synaptic weights at different layers, parameters in population encoder, and group decoder. Many candidate methods for tuning multi-layer SNNs have been proposed, including approximate BP \cite{cheng2020finite,zenke2018superspike}, equilibrium balancing \cite{RN762,RN761}, Hopfield-like tuning \cite{RN758}, and biologically plausible plasticity rules \cite{RN711}.

Here, we select the approximate BP for its efficiency and flexibility. The key feature of the approximate BP is replacing the non-differential parts of spiking neurons during BP to a predefined gradient number, shown as equation (\ref{equa_appro}), where we use the rectangular function equation to approximate the gradient of a spike.
\begin{equation}
    z(V) = \begin{cases}
        1\ \ & \text{if}\ |V-V_{th}| < w\\
        0\ \ & \text{otherwise}
    \end{cases}\text{,}
    \label{equa_appro}
\end{equation}
where $z$ is the pseudo-gradient, $V$ is membrane voltage, $V_{th}$ is the firing threshold and $w$ is the threshold window for passing the gradient.

\begin{figure}[htbp]
	\centering  
	\includegraphics[width=1.0\columnwidth]{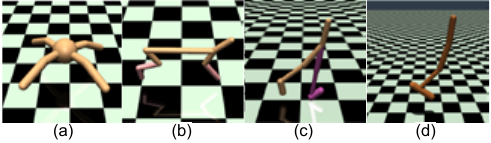}
	\caption{Four OpenAI gym tasks: (a) Ant-v3: make a four-legged creature walk forward as fast as possible, (b) HalfCheetah-v3: make a 2D cheetah robot run as fast as possible, (c) Walker2d-v3: make a 2D bipedal robot walk forward as fast as possible, and (d) Hopper-v3: make a 2D one-legged robot hop forward as fast as possible.}\label{fig-task}
\end{figure}

\section{Experiments}

To evaluate our model, we measured its performance on four continuous control tasks from the OpenAI gym (Fig. \ref{fig-task})~\cite{brockman2016openai}. 

\subsection{Implement Details}
Due to recent concerns in reproducibility~\cite{henderson2018deep}, all our experiments were reported over $10$ random seeds of the network initialization and gym simulator. Each task was run for $1$ million steps and evaluated every $10$k steps, where each evaluation reported the average reward over $10$ episodes without exploration noise, and each episode lasted for a maximum of $1000$ execution steps.

We compared our MDC-SAN against DAN and Pop-DAN (integrated population coding with DAN; it had the same amount of parameters as MDC-SAN for a fair comparison). 
DAN, Pop-DAN, and our MDC-SAN were all trained in conjunction with deep critic networks of the same structure using TD3 algorithm \cite{fujimoto2018addressing}. We evaluated the trained DAN, Pop-DAN, and MDC-SAN on four continuous control tasks (Fig. \ref{fig-task}) under the same settings and compared their performance (rewards gained). Pop-DAN and MDC-SAN used the same hyper-parameters as the DAN unless explicitly stated. The hyperparameter configurations of these models were as follows:
\paragraph{Integrate DAN with TD3} Actor network was (256, relu, 256, relu, action dim $M$, tanh); critic network was (256, relu, 256, relu, 1, linear); actor learning rate was $10^{-3}$; critic learning rate was $10^{-3}$; reward discount factor was $\gamma=0.99$; soft target update factor was $\eta=0.005$; maximum length of replay buffer was $T=10^6$; Gaussian exploration noise was $\sigma=0.1,\Tilde{\sigma}=0.2$; noise clip was $c=0.5$; mini-batch size was $n=100$; and policy delay factor was $d=2$;
\paragraph{Integrate Pop-DAN with TD3} Actor network was (Population Encoder, 256, relu, 256, relu, Group Decoder, action dim $M$, tanh); input population size for single state dimension was $J=10$; input coding used population coding ($C_{pop}$ for all tasks);
\paragraph{Integrate MDC-SAN with TD3} MDC-SAN used (Population Encoder, 256, DNs, 256, DNs, Group Decoder, action dim $M$, tanh), where the current decay factor and firing threshold of DNs were both $0.5$; input population size for single state dimension was $J=10$; MDC-SAN learning rate was $10^{-4}$; input coding used population coding ($C_{pop}$ for all tasks).

\begin{figure*}[t]
	\centering
	\includegraphics[width=0.95\textwidth]{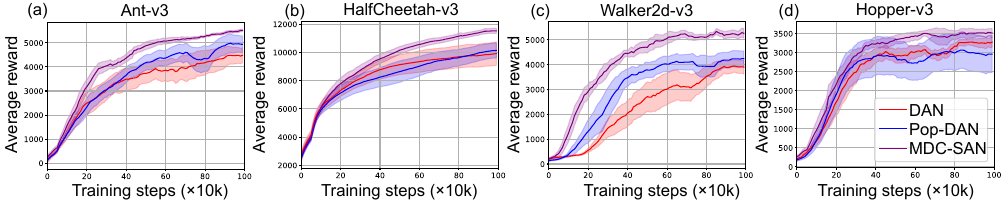}
	\caption{Comparison of average rewards for different models. (a) Performance of DAN, Pop-DAN, and MDC-SAN during training on the Ant-V3 gym task. (b, c, d) Performances of these three models on HalfCheetah-v3, Walker2d-v3, and Hopper-v3, respectively, where our proposed MDC-SAN performed best. The shaded region represents half a standard deviation of the average evaluation over 10 seeds, and the curves are smoothed for clarity.}\label{fig-main}
\end{figure*}

\begin{figure*}[htbp]
	\centering  
	\includegraphics[width=0.95\textwidth]{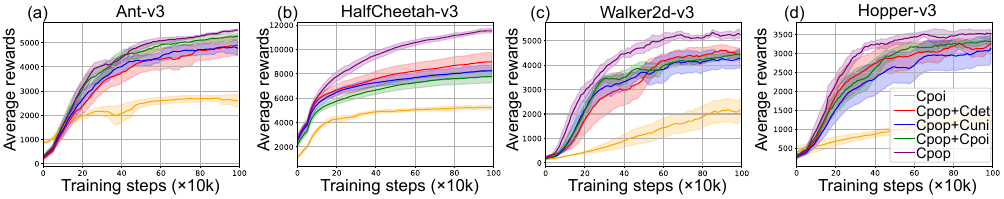}
	\caption{Comprehensive comparison of the impact of various input coding methods, where the SAN using $C_{pop}$ achieved the best performance.}\label{fig-code_type}
\end{figure*}

\subsection{Pre-learning of The Dynamic Parameters in DNs} \label{meta}

We selected Ant-v3 as the basic source task to pre-learn the dynamic parameters of DNs inside the MDC-SAN. All MDC-SAN parameters, including synaptic weights and dynamic parameters, were tuned with approximate BP. The learning curve (not shown) was continuously increased and converged after around 1 million training steps.

\begin{figure}[htbp]
	\centering  
	\includegraphics[width=1.0\columnwidth]{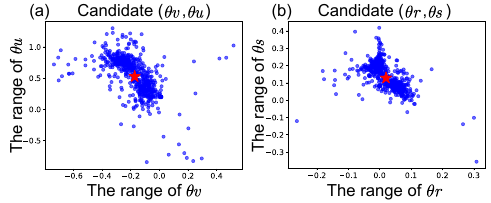}
	\caption{The clustering of the candidate dynamic parameters $\theta_{v,u,r,s}$ after learning on Ant-v3. (a) The parameters of $\theta_v$ and $\theta_u$ learned from Ant-v3. The single center of the cluster was measured by the standard k-means algorithm with k=1 (labeled as a red star). (b) Similar to that of (a) but for candidate parameters of $\theta_r$ and $\theta_s$.}\label{fig-learn_dns}
\end{figure}

After learning, all dynamic parameters related to DNs were clustered to ensure the optimal parameter configuration by selecting their clustering center. As shown in Fig. \ref{fig-learn_dns}(a, b), we obtained a clustering center of parameters $\theta_{v,u,r,s}$ (the red stars). Here we set $k=1$ in k-means for simplicity. The best dynamic parameters of DNs were $\theta_v^{*}$=-0.172 (conductivity of membrane potential), $\theta_u^*$=0.529 (conductivity of hidden state $U$), $\theta_r^*$=0.021 (reset membrane potential), and  $\theta_s^*$=0.132 (spike effect to $U$). These parameters were denoted as $\bm{\theta}^*$ and further used as the fixed configuration of all dynamic neurons for all tasks in the following experiments.

\begin{table}[htbp]
\centering
\fontsize{9}{10}\selectfont %

\begin{tabular}{@{}ccccc@{}}
\toprule
Actor Network & Ant        & HalfCheetah & Walker2d & Hopper \\ \midrule
DAN                  & 4671 &  10020       &       4139     &   3396      \\
Pop-DAN              &  5146  &    10206        &   4361           &     3201       \\
MDC-SAN              &  \textbf{5628}  &  \textbf{11657}              &     \textbf{5566}         &   \textbf{3636}          \\ \bottomrule
\end{tabular}
\caption{Max average rewards over 10 random seeds for DAN, Pop-DAN and MDC-SAN. Bold numbers are maximal values.}
\label{MAR}
\end{table}

\subsection{Benchmarking MDC-SAN against DAN}\label{main}

\begin{figure*}[t]
	\centering  
	\includegraphics[width=0.95\textwidth]{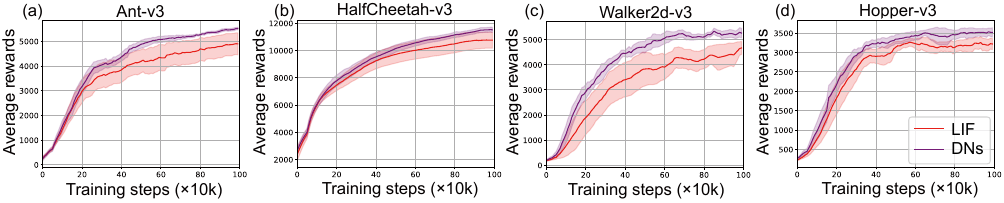}
	\caption{The DNs consistently performed better than LIF neurons on all four tasks.}
	\label{fig-dn-r-g}
\end{figure*}

We compared the performance of our MDC-SAN with DAN and Pop-DAN. As Fig. \ref{fig-main} and Table \ref{MAR} show, our model achieved the best performance across all four tasks, showing the effectiveness of our proposed MDC-SAN for continuous control tasks. In addition, Pop-DAN had no obvious advantage over DAN on the four tasks. Further analysis in Fig. \ref{fig-code_type} showed that SAN with population coding achieved significant performance improvements compared to that without population coding. SAN seems to be more compatible with 
population coding than DAN.

\subsection{Further Discussion of Different Input Codings}\label{coding}
We comprehensively compared the impact of various input coding methods on performance while keeping the neuronal coding method fixed to DNs. As Fig. \ref{fig-code_type} shows, performance of the rate coding method alone ($C_{poi}$, encode the input state with Poisson coding directly) was far inferior to population coding-based methods ($C_{pop}+C_{uni}$, $C_{pop}+C_{poi}$, $C_{pop}+C_{det}$, and $C_{pop}$) on all four tasks. This might be because the rate coding methods have an inherent limitation on the representation capacity of individual neurons. And population coding can better separate different states in a higher-dimension manner to produce better input representation. As for the population coding-based methods, $C_{pop}$ achieved the best performance on all tasks. And the other three population coding-based methods had little difference in performance. Hence, it seemed to be more efficient to directly use the analog value of the state after population coding as the network input, without further using rate coding to encode analog value into spike trains.

\subsection{The Neuronal Analysis in MDC-SAN}\label{MDN}

We further tested the constructed DNs on all four tasks and compared them with the LIF neurons while keeping the input coding method fixed to pure population coding ($C_{pop}$). As shown in Fig. \ref{fig-dn-r-g}, the DNs achieved better performance than LIF neurons on all tested tasks, including the source task (where the dynamic parameters of DNs were learned, i.e., Ant-v3) and other similar tasks (i.e., HalfCheetah-v3, Walker2d-v3, and Hopper-v3). This result verified the generalization capabilities of DNs, i.e., the dynamic parameters of DNs learned from a task could be generalized to other similar tasks. 

\begin{figure}[t]
	\centering  
	\includegraphics[width=1.0\columnwidth]{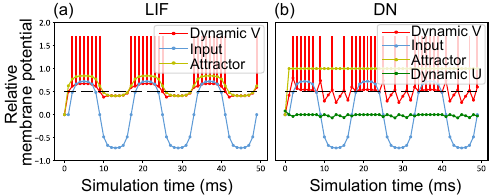}
	\caption{Membrane potential of DNs and LIF. The red lines represent dynamic membrane potential $V$, green lines are dynamic resistance item $U$, blue lines are simulated input, and yellow lines are the values of equilibrium points for membrane potentials.}\label{fig-lif-dn}
\end{figure}

One possible reason for the performance difference between LIF neurons and DNs was that the DNs contained a higher-order dynamics of membrane potentials, a more complex configuration of conductivity (both $\theta_v$ and $\theta_u$) and reset potential ($\theta_r$). Hence, the model using DNs showed a higher complexity at spatial-temporal information processing, which might contribute to the higher performance. An additional simulation of these two types of neurons was given, shown in Fig. \ref{fig-lif-dn}, including the neuronal dynamics for different explicit (e.g., membrane potential $V$ and stimulated input $I$) and implicit variables (e.g., resistance item $U$).

For the standard LIF neuron in Fig. \ref{fig-lif-dn}(a), the membrane potential was positively proportional to the neuron input. For example, for the sin-like input with the value range from -1 to 1, the dynamic $V$ was dynamically integrated until reaching the firing threshold $V_{th}$ only with a strong positive stimulus, or else decay accordingly with weak-positive or negative stimulus. Unlike LIF neurons, the DNs showed a higher complexity with an additional implicit $U$, making the dynamical changing of equilibrium points different. The slight differences of $U$ would cause a significant update of $V$ according to the definition of DNs, especially when the parameter $\theta_u$ is small in Equation (\ref{equa-MDN}). Hence, the DNs would show similar firing patterns with the strong positive stimulus and exhibit a sparse firing with the weak-positive or negative stimulus, instead of stopping firing like LIF neurons. This result showed a better dynamic representation of DNs than LIF neurons.

\section{Conclusion}

Efficient coding at multi scales is essential for the next-step decision-making in biological neural networks. This paper incorporates spatial population-coding at the input layer and temporal DN-coding at hidden layers as an integrative spiking actor network (MDC-SAN), reaching a better performance on the four benchmark Open-AI gym tasks than its counterpart DAN. Unlike generally used LIF neurons in SAN, the more complex DNs have shown a more vital ability on temporally non-linear information processing and achieved higher performance. In addition, population coding has also demonstrated an advantage in spatial information coding. 

In the future, more biologically plausible principles can be borrowed from biological networks and applied to SAN towards lower energy cost, more vital adaptability, and higher robust computation. These interactions between neuroscience and artificial intelligence have much in store for the future.

\section*{Acknowledgments}
This work was supported by the National Key R\&D Program of China (2020AAA0104305), the National Natural Science Foundation of China (61806195), the Strategic Priority Research Program of the Chinese Academy of Sciences (XDA27010404, XDB32070000).

\bibliography{aaai22}




\end{document}